\documentclass[10pt, conference, compsocconf]{IEEEtran}
\usepackage{url}

\usepackage{cite}

\ifCLASSINFOpdf
\else
\fi

\usepackage{graphicx}
\usepackage{lipsum}
\usepackage{booktabs}
\usepackage{threeparttable}
\usepackage{array}

\usepackage{xcolor}
\usepackage[ruled,vlined, linesnumbered]{algorithm2e}

\begin{document}
%
\title{PopEval: A Character-Level Approach to End-To-End Evaluation \\Compatible with Word-Level Benchmark Dataset}


\author{\IEEEauthorblockN{Hong-Seok Lee, Youngmin Yoon, Jang Pil Hoon, Chankyu Choi}
\IEEEauthorblockA{NAVER Corporation\\
$\{{hongs.lee }\cdot{youngmin.yoon}\cdot{ph.jang}\cdot{chankyu.choi}\}@navercorp.com$}}


%


\maketitle

\begin{abstract}
The most prevalent scope of interest for OCR applications used to be scanned documents, but it has now shifted towards the natural scene. Despite the change of times, the existing evaluation methods are still based on the old criteria suited better for the past interests. In this paper, we propose PopEval, a novel evaluation approach for the recent OCR interests. The new and past evaluation algorithms were compared through the results on various datasets and OCR models. Compared to the other evaluation methods, the proposed evaluation algorithm was closer to the human's qualitative evaluation than other existing methods. Although the evaluation algorithm was devised as a character-level approach, the comparative experiment revealed that PopEval is also compatible on existing benchmark datasets annotated at word-level. The proposed evaluation algorithm is not only applicable to current end-to-end
tasks, but also suggests a new direction to redesign the evaluation concept for further OCR researches.

\end{abstract}

\begin{IEEEkeywords}
end-to-end evaluation; character level evaluation; character-oriented evaluation, optical character recognition; 

\end{IEEEkeywords}

%
\IEEEpeerreviewmaketitle

\section{Introduction}
Evaluation metric is not merely a performance evaluation and ranking system of competition, but a navigator of optical character recognition(OCR) research because the direction of developing a certain model is heavily affected by its evaluation method.
Therefore, the evaluation metric should reflect actual performance of the model.
In this study, we investigated the theoretical bases of existing evaluation algorithms, and suggest a novel evaluation concept optimized to tasks of which current OCR researches are mainly focused: robust reading. 

In summary, our contributions are as follows: 1) We propose a novel character-oriented end-to-end evaluation protocol, compatible with existing benchmark datasets annotated at word level. 2) To confirm the compatibility between PopEval method and the word-level annotated benchmark datasets, we newly reannotated and published the most widely used test datasets for end-to-end system: focused scene text(ICDAR2013)  and incidental scene text(ICDAR2015) at character-level as quadrilaterals \cite{Karatzas2013ICDARCompetition, Karatzas2015ICDARReading}. 3) we performed the comparative analysis among evaluation metrics, detection-recognition algorithms and representative test datasets, then the results were compared with human qualitative end-to-end evaluation. 

The source code of the PopEval and the test datasets of ICDAR2013 and ICDAR2015 which were newly annotated at character-level are available at: \url{https://github.com/naver/popeval}

\section{Related works}
\subsection{Detection Evaluation}
In ICDAR2013 competition, DetEval was adopted as a detector evaluation metric at object level, that determines the matching objects by using double threshold system based on pixel precision and pixel recall \cite{Wolf2006ObjectAlgorithms}. DetEval also handles one-to-many(split) and many-to-one(merge) matching problems, but as it uniformly handles these match cases as same weight irrespective of the match condition, it results in errors. In addition, there have been similarity measuring methods to solve one-to-many and many-to-one problems, but these approaches required feature extraction \cite{khurshid2012word}. 

ICDAR2015 competition adopted the intersection over union(IOU) based PASCAL EVAL as an evaluation metric \cite{Everingham2015TheRetrospective}. If the IOU between two object areas exceeds 0.5, then the objects are considered as a match. In the IOU method, because a ground truth(GT) object only matches one predicted object, the split and merge problems are ignored \cite{Karatzas2015ICDARReading}.

COCO-Text competition adopted average precision(AP) with IOU \cite{Gomez2017ICDAR2017COCO-Text}. It required additional confidence rate values of detected objects to be calculated. The split and merge problems are not handled because it uses the concept of IOU.

\subsection{Recognition Evaluation}
For recognition tasks, total edit distance and correctly recognized words rate were adopted as the evaluation metric \cite{Karatzas2013ICDARCompetition,Karatzas2015ICDARReading}. The above performance indicator values have been calculated for both case sensitive and case insensitive. In correctly recognized words rate, one exactly matched recognized sequence is counted as one matching case regardless of the length of the transcript.

\subsection{End-to-End Evaluation}
Conventional end-to-end evaluation method is a pipeline that consists of detection and recognition phases.

\begin{figure*}[ht]
\centering
\includegraphics[width=\textwidth]{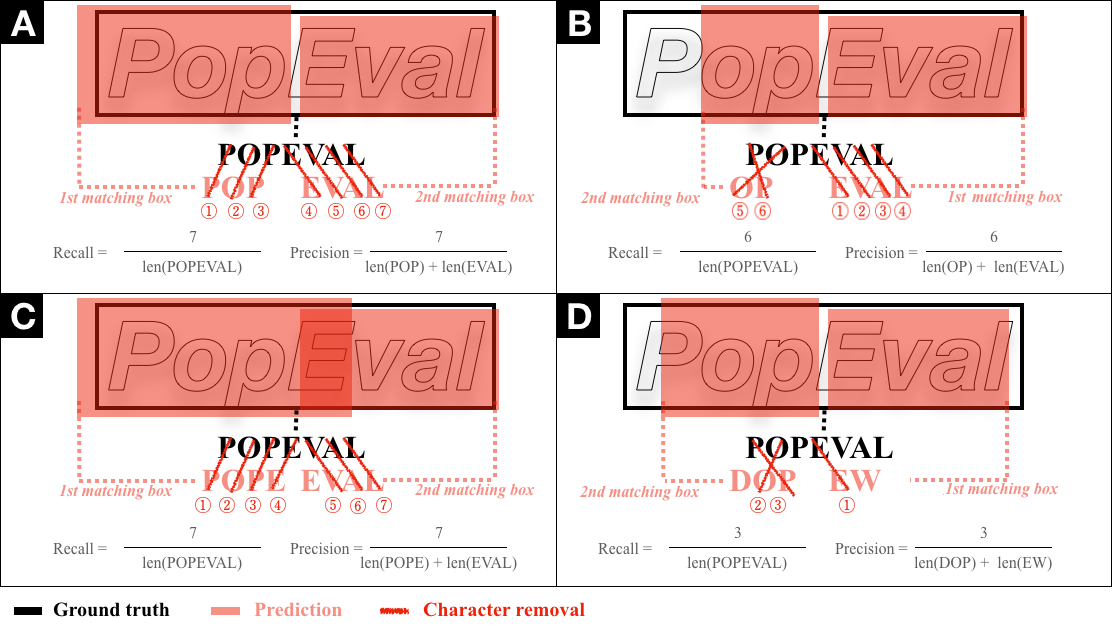}
\caption{PopEval character removal process scheme. The black box and text are GT and the red box and text are prediction. The order of character removal is indicated as circle character. (A) The GT polygon and text label `POPEVAL' and The predicted polygons and text labels `POP', and `EVAL'; (B) Deletion case: `OP' and 'EVAL' were predicted. There were six removed characters, one remaining character of GT; (C) Insertion case: `POPE' and `EVAL' were predicted. There were seven removed characters, and one remaining character of prediction; (D) Complicated case: `DOP' and `EW' were predicted. There were three removed characters, four remaining characters of GT, two remaining character of prediction; }
\label{fig1}
\end{figure*}

\subsubsection{Detection}
In ICDAR2013, ICDAR2015, and COCO-Text competition, the detected objects of which IOU is greater than 0.5 with the corresponding GT object passes to recognition phase. 
\subsubsection{Recognition}
In the ICDAR2013 and ICDAR2015 competitions, the recognition part adopted controlled vocabulary system which defines minimal common conditions to grant meaningful performance comparison.

COCO-Text competition does not use the vocabulary system, and it adopts the case-insensitive correctly recognized words method. By using the correctly recognized words method, the results passed through the detection phase are finally selected, then the corresponding APs are calculated using residues through the both phase\cite{Gomez2017ICDAR2017COCO-Text}. The AP has a drawback that is not intuitive to understand the absolute performance level of a certain model \cite{Moffat2008Rank-BiasedEffectiveness}.

Otherwise, the normalized edit distance(1-NED) can be used. After the detection phase, the edit distance of the GT and the recognized transcript is divided into the most long length of the GT or the recognized transcripts then subtract the average results to 1. If the detected box is not caught in the detection phase, the calculation of 1-NED is performed by assuming that the recognized transcripts were blank \cite{Shi2017ICDAR2017RCTW-17}.

\section{PopEval: Our approach}

PopEval is a character level end-to-end evaluation metric that is based on removing overlapping characters between the GT and the OCR result. 
In this criteria, the number of removed characters is considered as true positive count to be used for calculating character recall and precision. 
Contrary to existing end-to-end evaluation metrics, PopEval is not just a pipeline structure which consists of detection and recognition parts, but a seamless structure of which character removal is conducted by integration of detection and recognition results.  

The principle of PopEval algorithm is to adopt how human beings evaluate a certain recognized result comparing with GT. 
Based on the principle, there are three criteria in which characters are removed.

\begin{algorithm}[t]
    \SetKwFunction{hi}{hi}
     
    \definecolor{mygray}{gray}{0.6}
    \SetKwFor{For}{for}{in \textit{one-to-one-relations}}{end}
    \SetKwFor{Forchar}{for}{in \textit{Det\_text}}{end}
    
    \SetKwFor{ForGT}{for}{in \textit{GTs}}{end}
    
    \SetKwFor{ForOneToMany}{for}{in \textit{one-to-many}}{end}
    
    \SetKwProg{ifprog}{if}{}{end}
    \SetKwRepeat{repeat}{a}{b}

    \SetKwProg{Def}{def}{}{end}
    \SetAlgoLined
    \textbf{global} removed\_char\_count = 0\\
     \Def{main(GTs, Dets):}{
        \textit{one-to-one}, \textit{one-to-many} = \textit{InspectRelation(GTs, Dets)}\\
        \textcolor{mygray}{
            \textit{\# \textit{one-to-one} = [[GT1, Det1], [GT2, Det2], ...]}\\
            \textit{\# \textit{one-to-many} = [[GT3, [Det3, Det4]], ...]}\\
        }
        \textbf{if} \textit{len(one-to-one)} is 0 \textbf{and} \textit{len(one-to-many)} is 0\\
        \ \ \ \ \textbf{return} \textit{removed\_char\_count}\\
        \textbf{else if} \textit{len(one-to-one)} is 0\\
        \ \ \ \ \textbf{return} \textit{HandleOneToMany}(\textit{one-to-many})\\
        \textbf{else}\\ 
        \ \ \ \ \textit{remaining\_GTs}, \textit{remaining\_Dets} = \textit{CharacterRemovalProcess}(one-to-one)\\
        \ \ \ \ \textbf{return} \textit{main}(\textit{remaining\_GTs}, \textit{remaining\_Dets})\\
        \textbf{end}
     
     }
     
     \Def{InspectRelation(GTs, Dets):}{
        initialize \textit{one-to-one}, \textit{one-to-many} as array
        
        \ForGT{GT}{
            \textit{Dets\_intersect} = filtered \textit{Dets} intersecting \textit{GT}\\
            \textbf{if} \textit{len(\textit{Dets\_intersect})} is 1\\
            \ \ \ \ \textit{one-to-one}.append( [\textit{GT}, \textit{Dets\_intersect}] )\\
            \textbf{else if} \textit{len(\textit{Dets\_intersect})} over 1:\\
            \ \ \ \ \textit{one-to-many}.append( [\textit{GT}, \textit{Dets\_intersect}] )\\
            \textbf{end}
        }
        \textbf{return} \textit{one-to-one}, \textit{one-to-many}
     }
     
     \Def{HandleOneToMany(\textit{one-to-many}):}{
        \textit{GTs}, \textit{Dets} = export \textit{GTs} and \textit{Dets} from \textit{one-to-many}\\
        \textit{GT} = closest one to left top of image among GTs \\
        \textit{Det} = one with the highest area recall on \textit{GT} among \textit{Dets}\\
        \textit{remaining\_GTs}, \textit{remaining\_Dets} = \textit{CharacterRemovalProcess}(\textit{GT, Det, GTs, Dets})\\
        \textbf{return} \textit{Main}(\textit{remaining\_GTs}, \textit{remaining\_Dets})
     }
    \caption{PopEval evaluation metric, part 1 of 2}
\end{algorithm}

\subsection{Determining the Iteration Order of GT Polygons} 
People read English text as images, from top left to bottom right. Likewise, the iteration order of scattered GT on each image is determined as the distance between the GT polygon centroid and left-top point of the image. The shorter the distance, the earlier the iteration sequence order. This approach solves many-to-one problems in relation to GT-to-detection. If there are multiple GTs corresponding to one detection, then the iteration order is predetermined. 
\begin{equation}
\newcommand{\arga}{arg}
\newcommand{\mina}{min}
\newcommand{\argmin}{\arg\!\min}
k \in {\{ 1, .. , n\}}
\end{equation}
\begin{equation}
\newcommand{\arga}{arg}
\newcommand{\argmin}{\arg\!\min}
{f(k)} = \sqrt{{X_{k}}^2 + {Y_{k}}^2} 
\end{equation}
\begin{equation}
\newcommand{\argmin}{\arg\!\min}
D = {\argmin_x} {f(x)} 
\end{equation}

For each iteration, the index of remaining GT polygons is $k$. The coordinates of $k$-th polygon centroid are $X_k$, $Y_k$.
The distance between origin and the $k$-th polygon centroid is $f$. $D$ indicates the nearest polygon to left-top of the image, that is subjected to next processes.

\begin{algorithm}[t]
\setcounter{AlgoLine}{33}

\SetKwFunction{hi}{hi}
 
\definecolor{mygray}{gray}{0.6}
\SetKwFor{For}{for}{in \textit{one-to-one-relations}}{end}
\SetKwFor{Forchar}{for}{in \textit{Det\_text}}{end}

\SetKwFor{ForGT}{for}{in \textit{GTs}}{end}

\SetKwFor{ForOneToMany}{for}{in \textit{one-to-many}}{end}

\SetKwProg{ifprog}{if}{}{end}
\SetKwRepeat{repeat}{a}{b}

\SetKwProg{Def}{def}{}{end}
\SetAlgoLined
\Def{CharacterRemovalProcess(\textit{GT, Det, GTs, Dets}):}{
    \textit{GT\_text}, \textit{Det\_text} = Extract texts from \textit{GT}, \textit{Det}\\
    
    \Forchar{\textit{Det\_character, Detchar\_index}} {

        \ifprog{ \textit{Det\_character} in \textit{GT\_text}}{
            \textit{GTchar\_index} = the leftmost character index of \textit{GT\_text} matching  \textit{Det\_character}
            
            \textbf{delete} a character of \textit{GTchar\_index} from \textit{GT\_text}\\
            \textit{removed\_char\_count} += $1$    \\
            \ifprog{GT\_text is empty}{ \textbf{delete} \textit{GT} object from \textit{GTs} \\

            }
        }

        \ifprog{\textit{Det\_character} is last of \textit{Det\_text}}{ \textbf{delete} \textit{Det} object from \textit{Dets} }
    }
    \textit{GTs}, \textit{Dets} = the deleted \textit{GT} and \textit{Det} objects should be excluded from \textit{GTs}, \textit{Dets}\\
    \textbf{return} \textit{GTs}, \textit{Dets}

 }
 \caption{PopEval evaluation metric, part 2 of 2}
\end{algorithm}

\subsection{Handling the One-to-Many Relations} \label{handleonetomany}
When the one-to-many relations are encountered in the course of GT-to-detection, it is required to pick one of the detected polygons matching the GT polygon in one-to-one relation. 
To solve this problem, PopEval adopted a recursive procedure which repeats inspecting every relation among GT polygons and detected polygons then preferentially taking out the obvious one-to-one relations, until only one-to-many relations remain. 
While the recursion is repeatedly executed, the previous one-to-many relations can be converted to one-to-one relationship in the current operation because obvious relations were wiped out in previous execution. 
For the last remaining one-to-many relations, the area recall of each detected polygon is adopted as the determinant to select one detected polygon that is matched with the GT: the larger the area recall, the higher the priority. 
Generally, only one detected polygon is subjected to character removal process.  However, when there are multiple detected polygons with the same highest area recall, these are subjected to the process while being weighted as a reciprocal of the number of the subjected detections. 
\subsection{Character Removal Process} \label{crp}

For character removal, the basic unit of comparison is a set of polygon area and the transcript. PopEval compares the units among GTs and predicted results. If the polygon areas of GT and predicted result overlap to each other, the transcriptions of GT and predicted result are compared, then the overlapped character is removed one by one in the predetermined order. 

The principle of PopEval algorithm is to adopt the way a human evaluates. In the principle, there are two rules about the order in which characters are removed.

First, in the unit, the removal iteration of characters is conducted in the direction how characters are read. In this study, as the test dataset was in English, the iteration order of character removal was left to right. For (B) in Figure \ref{fig1}, the GT transcript is ``POPEVAL'', and the recognized transcripts are ``OP'', and ``EVAL''. According to \textit{handling the one-to-many relations}(\ref{handleonetomany}), a unit of ``EVAL'' is subjected to the \textit{character removal process}(\ref{crp}) first. The transcript of GT is removed from the characters of ``EVAL'' in order from left to right, then the transcript of GT becomes ``POP''. 

Second, if a character of recognized transcript corresponds to multiple characters in the GT, the criteria to remove one character stays the same as above, following the direction in which characters are read in the language. Continuing from the above example where the transcript of GT became ``POP'', and last recognized transcript was ``OP''. According the character removal order, ``O'' is removed first, then ``P'' will drop out of transcripts. However, when ``P'' is removed, there are two candidates of character removal in GT transcript ``POP'', the first and third. As the direction determined above,
the first character of the ``POP'' is picked to be matched and removed together with ``P'' of recognized transcript.
As final result, the remaining GT transcript is ``P'' and there is no remains in recognized transcripts.

Recall and precision are calculated from the lengths of the initial GT and recognized transcript, and the number of removed characters. In this example, the length of the initial GT is seven, the initial total length of recognized transcript is six, and six characters were removed. The precision and recall are 1.0 and 0.8571, respectively.

\section{Experimental Result}
\subsection{Inspection on the Case of Concern in the PopEval}
Currently, there is no perfect metric in the evaluation of OCR \cite{Long2018SceneEra}, and it is important that which metric is actually more accurate.
Since PopEval is a method to remove overlapped character components between objects, there is a room for concern that it may not reflect the permutation problem in which the recognized transcript has different character arrangement compared to GT transcript.
Therefore, the permutation problem was monitored by inspecting how frequently the problem occurs on state-of-the-art recognition models: attentional scene text recognizer with flexible rectification(ASTER) \cite{Shi2018ASTER:Rectification}, and gated recurrent convolution neural network for OCR(GRCNN) \cite{Wang2017GatedOCR}.
Table 1 shows the occurrence of permutation problems on recognition models and test datasets. Test datasets of ICDAR2013 and ICDAR2015 were inspected.
The permutation problem is defined as below:\\\\
\textit{ 1) The transcripts of GT and recognition have the same character component.}\\
\textit{ 2) The character arrangements of the transcripts are different to each other.}\\

The survey showed that the permutation problem has rarely occurred. In the case review of the results, it is found that common permutation occurrences on the two models were both caused by a typing error in the GT. Subsequently, there was no permutation problem in test dataset of ICDAR2013 and out of 1811 images in total, the permutation occurred once with ASTER model, twice with GRCNN model for the dataset of ICDAR2015. 
Therefore, the occurrence of permutation problem is rare, considered as scarcely impinge on evaluation. 

\begin{table}[!t]
\caption{Among the recognition results which composed of the same alphanumeric components as GT, the proportion that does not exactly match GT.}
\label{table_example}
\centering
\begin{tabular}{l c c}
\hline
 & ICDAR2013 & ICDAR2015 \\ \hline
ASTER & 0.00\% & 0.05\% \\ 
GRCNN & 0.00\% & 0.14\% \\ \hline
\end{tabular}
\end{table}
\subsection{Occurrence of One-to-Many and Many-to-One Relations}
IOU thresholding and exact text matching methods only accept one-to-one relations \cite{Long2018SceneEra}. To inspect the errors caused by ignoring one-to-many and many-to-one relations, PixelLink\cite{Deng2018PixelLink:Segmentation} and EAST\cite{ZhouEAST:Detector} were adopted as the detection model, and the ASTER and the GRCNN recognition models were trained as recognition model and made to predict on the test datasets of ICDAR2013 and ICDAR2015 competition.

One-to-many and many-to-one relations were counted on Table II under below criteria . 
\subsubsection{One-to-Many(split)}
If the recognized transcript of either GRCNN and ASTER is included in the GT transcript and the area precision of detection and GT boxes is greater than 0.5, the detection box is counted as a box in one-to-many relation.

\subsubsection{Many-to-One(merge)}
If a GT transcript is a part of the recognized transcript of either GRCNN and ASTER and the area recall of detection and GT boxes is greater than 0.5, the GT box is counted as a box in many-to-one relation.

The assessment found that a non-negligible number of detection boxes and GT boxes were in one-to-many and many-to-one relations. Although there is ambiguity that the boxes in the relations match well each other in terms of shape and area, the transcript of the boxes is still valuable. In the approach ignoring these relations, all of the split detections and merged GTs are evaluated as false negatives. Additionally, since the relation assessment aforementioned relies on an imperfect recognition model, it is expected that there will be more cases of one-to-many or many-to-one relations than the occurrences assessed.
\begin{table}[!t]
\caption{The proportions of split detections and merged GTs among the total detections and GTs, respectively.}
\label{table2}
\centering
\begin{tabular}{ccc}
\hline
 & Split Detections & Merged GTs \\ 
 & (one-to-many) & (many-to-one) \\ \hline
EAST - ICDAR2013 & 3.84\% & 1.46\% \\ 
PIXEL - ICDAR2013 & 6.09\% & 3.29\% \\
EAST - ICDAR2015 & 1.13\% & 1.54\% \\ 
PIXEL - ICDAR2015 & 2.05\% & 0.35\% \\ \hline
\end{tabular}
\end{table}

\subsection{PopEval's compatibility with benchmark datasets annotated at word-level and character-level}
Since PopEval is an approach to evaluate the matched character component between GT and detection, it is most accurate when it is used on a character-level annotated benchmark dataset. Because benchmark datasets commonly have been annotated on word-level, we reannotated the test datasets of the ICDAR2013 and ICDAR2015 competitions on character level. 

OCR models were evaluated at word-level and character-level then the compatibility between the results at character and word levels was investigated. 
Efficient and accurate scene text detector(EAST) and PixelLink were adopted as the detector model and ASTER and GRCNN were adopted as the recognition model. Therefore, the four detector-recognizer models were established, then evaluated on each of word-level and character-level benchmark datasets. As a F1 score, the harmonic means of recall and precision were calculated on Table \ref{table3}. 
\begin{table}[!t]
\caption{Comparative analysis of PopEval F-score for word level and character level annotation of benchmark datasets.
}
\label{table3}
\centering
\begin{tabular}{l c c c}
\multicolumn{4} {l} {For ICDAR2013 Test Dataset} \\\hline
 & Word Level & Character Level & Diff\\ \hline 
EAST - ASTER & 0.8649 & 0.8616 & 0.0033\\ 
PIXEL - GRCNN & 0.8562 & 0.8531 & 0.0031 \\ 
EAST - ASTER & 0.8540 & 0.8513 & 0.0027\\ 
PIXEL - GRCNN & 0.8552 & 0.8538 & 0.0014 \\  \hline\\
\multicolumn{4} {l} {For ICDAR2015 Test Dataset} \\\hline
 & Word Level & Character Level & Diff\\ \hline 
EAST - ASTER & 0.8017 & 0.7991 & 0.0026\\ 
PIXEL - GRCNN & 0.7696 & 0.7661 & 0.0035 \\ 
EAST - ASTER & 0.7792 & 0.7783 & 0.0009\\ 
PIXEL - GRCNN & 0.8003 & 0.7986 & 0.0017 \\ \hline 
\end{tabular}
\end{table}
The difference of F1 score between the evaluations on word-level and character-level datasets constantly stayed below 0.004. Considering the minor difference between the evaluations on word-level and character-level annotatated datasets, therefore, PopEval is compatible with the existing benchmark datasets which were annotated at word-level .

\subsection{Correlations Between the End-to-End Evaluation Algorithms and Manual Qualitative Evaluation.}
Since each existing evaluation algorithm has its own limitation, it is difficult to quantitatively determine which algorithm is more accurate. In this study, a qualitative evaluation was manually performed as a standard to compare the evaluation algorithms. For the qualitative evaluation, the participants used an assistant tool visualizing locations and transcriptions of GT and OCR result. Considering ``do not care'' marking of ICDAR2015 \cite{Karatzas2015ICDARReading}, the predicted polygons corresponding to the ``do not care'' markings were removed as preprocessing of the qualitative evaluation. The performance was evaluated as a character-oriented method by considering the errors of insertion, deletion, and substitution. A percentile of performance was marked as a five point scale: 0\% to 20\%, 1 point; 20\% to 40\%, 2 point; 40\% to 60\%, 3 point; 60\% to 80\%, 4 point; and 80\% to 100\%, 5 point; 

To assess the correlation between evaluation algorithms and the manual qualitative evaluation, the average of three participants' scores and the results of following end-to-end evaluation algorithms were subjected to the assessment: the vocabulary-aided transcript matching with IOU over 0.5; the average precision with IOU over 0.5; the 1-NED; the PopEval using word-level dataset; and the PopEval using character-level dataset;
For OCR model subjected to the assessment, because the Pixelink obtains an object by postprocessing, there is an ambiguity in calculating the confidence rate of the object for measuring average precision(AP). Therefore, the EAST as a detection model and both of the recognition models were subjected to the assessment, then there were two OCR models to be evaluated. For benchmark dataset, the test datasets of ICDAR2013 and ICDAR2015 were subjected to the assessment.

Pearson correlation was adopted to assess linear correlations between the manual qualitative evaluation and the end-to-end evaluation algorithms. As the result of the assessment, the PopEval was found to be the most similar to the manual qualitative evaluation in all cases. For ICDAR2013, the PopEval with character-level dataset showed very high correlation as $0.946$ with the manual qualitative evaluation, nearly followed by the PopEval with word-level dataset. In Pearson correlation, coefficient above 0.8 means strong linear correlation in general.  Although the correlation between PopEval and manual evaluation relatively decreased for EAST-GRCNN model with ICDAR2015, it still showed a strong correlation with the manual evaluation, followed by the other algorithms, and the traditional algorithms also showed lower correlation for EAST-GRCNN model than for the other.  

This experiment showed that PopEval is the most correlated evaluation method with human qualitative evaluation among existing evaluation methods. Among traditional evaluation algorithms, the 1-NED showed the most correlation with manual qualitative evaluation.
\begin{table}[!t]
\caption{The Pearson correlation coefficients between the manual qualitative evaluation and the end-to-end evaluation algorithms for two OCR model.}
\label{table_example}
\centering
\begin{threeparttable}

\begin{tabular}{m{3.5em}m{3.7em}m{2.3em}m{4em}m{3.5em}m{4.5em}}

\multicolumn{6}{c}{For ICDAR2013 Test Dataset}\\
\hline
 & \centering\arraybackslash Vocab &\centering\arraybackslash AP &\centering\arraybackslash 1-NED &\centering\arraybackslash PopEval at word &\centering\arraybackslash PopEval at character\\
 \hline
EAST - ASTER &\centering\arraybackslash 0.7858 &\centering\arraybackslash 0.4595 &\centering\arraybackslash 0.8884 &\centering\arraybackslash 0.9305 &\centering\arraybackslash 0.9340 \\
EAST - GRCNN &\centering\arraybackslash 0.7910 &\centering\arraybackslash 0.4437 &\centering\arraybackslash 0.8800 &\centering\arraybackslash 0.9457 &\centering\arraybackslash 0.9461 \\
\hline
\\

\multicolumn{6}{c}{For ICDAR2015 Test Dataset}\\

\hline
 & \centering\arraybackslash Vocab &\centering\arraybackslash AP &\centering\arraybackslash 1-NED &\centering\arraybackslash PopEval at word &\centering\arraybackslash PopEval at character\\
\hline
EAST - ASTER &\centering\arraybackslash 0.7776 &\centering\arraybackslash 0.5792 &\centering\arraybackslash 0.8124 &\centering\arraybackslash 0.9272 &\centering\arraybackslash 0.9213 \\
EAST - GRCNN &\centering\arraybackslash 0.6870 &\centering\arraybackslash 0.5410 &\centering\arraybackslash 0.7262 &\centering\arraybackslash 0.8221 &\centering\arraybackslash 0.8204 \\
\hline

\end{tabular}

\begin{tablenotes}[para,flushleft]
    \footnotesize
    Vocab: vocabulary-aided transcript matching with IOU over 0.5; AP: average precision; 1-NED: normalized edit distance; PopEval at word: PopEval with word level dataset; PopEval at character: PopEval with character level dataset;
\end{tablenotes}
\end{threeparttable}

\end{table}

\section{Discussion and Conclusion}
As referred by Wolf and Jolion in \cite{Wolf2006ObjectAlgorithms}, the drawback of object-oriented matching is the requirement that the bounding box wraps the actual text area tightly. For this reason, the rectangle approach scheme was only suitable for document images. In contrast to document scanning, however, extracting text from natural scene is much more difficult as the texts come with many varieties such as different orientations, varying aspect ratios or even skewed shapes. To account for these varieties, the four-vertices polygon approach was adopted as an annotation method recently. For recent interests of OCR, however, the four-vertices polygon method has the same limitation that the text area should be wrapped tightly, especially for curved texts. Therefore, a new annotation method with polygons of unlimited number of vertices is needed.

In benchmark dataset with polygon of unlimited vertices, the conventional approach is not appropriate, such as IOU, DetEval and average precision at a specific IOU. For GT of quadrilaterals, most of the relations between GT and detection were one-to-one, and the conventional criteria concepts considered only one-to-one relations ignoring the others. However, because the split and merge relation occurs more frequently with datasets with polygons of unlimited vertices such like Total-Text dataset \cite{Chng2017Total-Text:Recognition}, the concept of the object matching should be changed to reflect the actual performance. 

In recognition, the vocabulary based evaluations are not adequate for wild scene text. Because the wild scene has varying texts  such as unique nouns \cite{Long2018SceneEra}, dictionary based end-to-end evaluation is by its structure incapable of handling wild scene text. Even in strongly and weakly contextualised evaluations \cite{Karatzas2015ICDARReading}, the dictionary based evaluation has a limitation of not reflecting actual performance. Therefore, the current evaluations of recognition are based on edit distance and exact matching method.  

When a recognition model recognizes a long string correctly, the model should be rated better than other models that recognized short strings. However, the exact matching method has its own drawback of not considering the various difficulty of each recognition because the method does not take into account partial correctness. The exact matching method causes underestimation of the model's actual performance, and the miscalculation depends on characteristics of benchmark dataset in use. Considering the above drawbacks, it is deemed desirable to approach the character-oriented evaluation rather than the object-oriented evaluation. Because the current OCR interests, such as multi-language transcripts, are more difficult to detect and recognize correctly than the previous tasks, the character-oriented evaluation is essential to evaluate the actual performance. In this aspect, the 1-NED was suggested as an end-to-end evaluation \cite{Shi2017ICDAR2017RCTW-17}. However, because it adopted IOU threshold as the criteria of object detection, this caused limitations due to threshold and ignoring split and merge relations.

Because the character-oriented evaluation requires character-level annotated dataset for accuracy evaluation, the character-level annotation should be provided as a test dataset in the future. 
Correspondingly, in order to develop PopEval, we newly annotated the existing benchmark datasets at character level.
Although PopEval was devised to evaluate benchmark datasets annotated at character level, the evaluation method can be applicable to word-level benchmark dataset. The experimental results show that PopEval is compatible with word-level annotation, meaning PopEval can evaluate previous end-to-end tasks at character-oriented level without re-annotating the datasets at word-level.

The PopEval is a consistent performance evaluator for various benchmark datasets. In benchmark datasets annotated as unoriented rectangle box, the texts were not tightly wrapped by the ground truth annotations. This ambiguity necessitates conventional evaluation metrics to use variable thresholds for different benchmark datasets \cite{Wolf2006ObjectAlgorithms}. Different thresholds need to be applied to different benchmark datasets based on their characteristics, and incorrect results can be occurred in this process \cite{Long2018SceneEra}. On the other hand, PopEval does not use the threshold method, but adopts pixel recalls between a GT and each detection, and this enhances the consistency of PopEval for various benchmark datasets. 

The PopEval is the most human-like end-to-end evaluation method. Although the concept of the edit distance has been an effective method for recognition evaluation, in the aspect of end-to-end evaluation, the 1-NED contains the incomplete detection evaluation criteria caused by IOU concept. 
Through the correlation assessment between human qualitative evaluation and the algorithms, the PopEval showed much higher correlation with the human evaluation than the 1-NED. It means PopEval can handle the imperfection case of 1-NED, making its results more similar to those done with human evaluation.

\begin{figure}[!t]
\centering
\includegraphics[width=7cm]{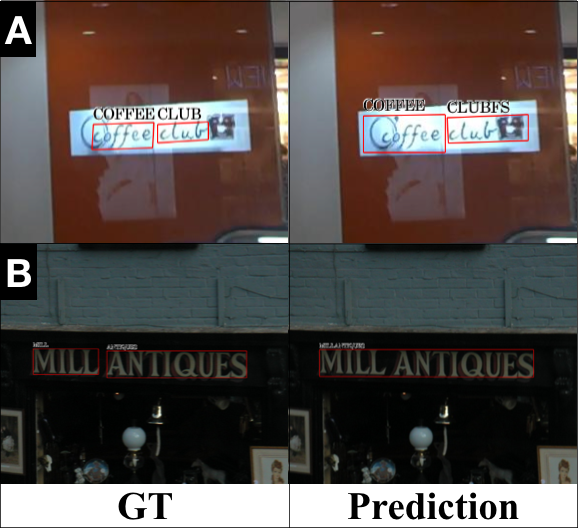}
\caption{The representative evaluation cases showing limitation of traditional evaluation methods. The followings are evaluation results of each image; A: the IOU threshold can not catch the detected objects. 0.9090(PopEval), 0.0(1-NED), 0.0(AP); B: the merge relation occurred.  1.0(PopEval), 0.33(1-NED), 0.0(AP);    }
\vspace{-0.2in}

\label{fig_sim2}
\end{figure}

The conventional evaluation methods such as DetEval, criteria of IOU and the edit distance have been adopted as the evaluation standard for a long time. Recently, it has been necessary to optimize the conceptual criteria of evaluation for new challenging tasks of OCR. In contrast to previous evaluation methods that are based on quadrilaterals, PopEval is able to handle polygons consisting of unlimited number of vertices. Above of all, the most innovative aspect is performing character-oriented evaluation with existing benchmark datasets annotated at word-level. 

Further study and experiments are expected to enhance the integrity of PopEval. As with the imperfection of the existing evaluation methods, the permutation problem is a point of concern in PopEval. Although the experiment showed that the permutation problem rarely occurred, it is expected that concepts like the n-gram of BLEU can be applied to handle the sequence of characters in further study \cite{Papineni2001BLEU}. The character removal method, which provides compatibility with word-level datasets, is expected to contribute to more accurate model performance evaluation for future OCR tasks.






%

\bibliography{references.bib}
\bibliographystyle{ieeetr}


\end{document}